\documentclass[sigconf]{acmart}

\usepackage{booktabs} 
\usepackage{mathtools}
\usepackage{amsmath}
\usepackage{balance}

\setcopyright{rightsretained}

\acmDOI{10.475/123_4}

\acmISBN{123-4567-24-567/08/06}

\acmConference[WWW 2018]{The Web Conference}{April 2018}{Lyon, France}
\acmYear{2018}
\copyrightyear{2017}
\acmPrice{15.00}

\begin{document}
\title{Find The Conversation Killers: A Predictive Study of Thread-ending Posts}
\author{Yunhao Jiao}
\affiliation{%
  \institution{Zhejiang University}
}
\email{jiao_yunhao@zju.edu.cn}

\author{Cheng Li}
\affiliation{%
  \institution{University of Michigan}
}
\email{lichengz@umich.edu}

\author{Fei Wu}
\affiliation{%
  \institution{Zhejiang University}
}
\email{wufei@zju.edu.cn}

\author{Qiaozhu Mei}
\affiliation{%
  \institution{University of Michigan}
}
\email{qmei@umich.edu}

\begin{abstract}
How to improve the quality of conversations in online communities has attracted considerable attention recently. Having engaged, urbane, and reactive online conversations has a critical effect on the social life of Internet users. In this study, we are particularly interested in identifying a post in a multi-party conversation that is unlikely to be further replied to, which therefore kills that thread of the conversation. For this purpose, we propose a deep learning model called the ConverNet. ConverNet is attractive due to its capability of modeling the internal structure of a long conversation and its appropriate encoding of the contextual information of the conversation, through effective integration of attention mechanisms. Empirical experiments on real-world datasets demonstrate the effectiveness of the proposal model. For the widely concerned topic, our analysis also offers implications for improving the quality and user experience of online conversations.
\end{abstract}




\maketitle

\section{Introduction}
\footnote{This work was done when the first author was visiting the University of Michigan.}More and more people are relying on online communities to access the latest information, exchange ideas, express options, and participate in discussions. Facilitating these natural conversations in online communities has become increasingly important. On one hand, decision makers utilize these conversations to optimize their online marketing strategies; social scientists study how opinions are shaped and diffused through discussions; politicians analyze how users respond to certain governmental policies. On the other hand, effective and healthy conversations lead to increasing satisfaction and engagement of users; low-quality and ill conversations hurt the users' social experience, turn them away, and even convert them into trolls. How to engage people in online conversations has aroused the interest of researchers from various domains.

Facilitating user conversations in online communities has also become an active line of research in the field of data mining. Many studies focus on predicting the number of retweets accumulated by a particular Tweet \cite{Suh2010, Hong2011, tan2014effect} in Twitter, identifying various factors that could help users get a higher response rate from the audience. Similar studies have also been conducted on online forums \cite{Wang2011, Balali}, e.g., to parse and predict thread structures, which can potentially enhance information access and support sharing.

While most existing work on online conversations focuses on posts that are positively received (e.g., highly retweeted Tweets), there are much fewer studies on posts that are negatively received. For example, there is no existing work identifying conversation killers, posts that result in no further replies in a multi-party conversation. 
Analyzing the ineffectiveness of conversations is as important as analyzing the effectiveness. Indeed, a conversation killer not only prevents new information from being introduced and new opinions from being expressed, but also projects negative experience to the author himself - the lack of response often leads to disappointment and lower self-evaluation, and further decreases their interest and engagement. Developing a system that identifies potential conversation-killing posts and providing suggestions accordingly could greatly improve the engagement of users into the conversations. For example, if a user intends to expand a discussion, the system could send out a notice before they submit their post if it is more likely to end the discussion instead. Such a notice, together with possible suggestions, is also plausible in a two-way conversation when one intends to have the other engaged. 

In this work, we study the novel task of predicting thread-ending posts, which we use as a practical surrogate for ``conversation killers.'' Although not all thread-ending posts are killing a conversation, and not all are done in an unintended way, knowing whether or not there will be further relies does help one to avoid becoming a potential conversation killer. We analyze various properties that are potentially predictive to ending a conversation, including text content, conversation background, conversation structure, and sentiments. We find that a standard SVM model is able to distinguish predictive signals from others. To make the best use of these predictive signals, a specially designed recurrent neural network (RNN), named the ConverNet, is employed to model conversations as sequences of posts, as RNNs are known to be good at learning high-level representations from the text. 
A particular challenge of modeling conversations is the large variance of the length of threads, which makes standard RNNs ineffective due to their weakness to handle long term dependency and standard attention mechanisms ineffective due to the inability of handling various-lengthed sequences. 
To address this challenge, we propose a simple yet powerful attention mechanism that is specifically designed for this task. The attention mechanism not only resolves the issue of lengthy threads, but also provides an effective way to model important context information (e.g., timestamps and authorship) in the conversation. 

We conduct large-scale experiments with conversations in two representative domains -- online forums (e.g., Reddit) and movie dialogs. The results demonstrate great effectiveness and generality of ConverNet, which outperforms a portfolio of strong baselines, including feature based SVMs and deep learning methods equipped with standard attention mechanisms. By comparing the results of ConverNet and SVM, we present interesting implications of how to engage the participants in a conversation.

\section{RELATED WORK}
The large quantity of information on the social media platforms exhibits a great potential for the research in the domain of data mining and natural language processing. In this paper, we specifically focus on the task of predicting posts that will be conversation endings. There are several lines of related work.

\subsection{Prediction of Replies or Retweets for Microblogs}
One line of work concerning response prediction on social networks aims to predict the number of replies. This helps content generators, especially advertisers and celebrities, to increase their exposure and maintain their images. 
Rowe et al. (2011) \cite{Rowe2011} targeted the prediction of seed posts and their potential number of replies in the future. Suh et al. (2010) \cite{Suh2010} and Hong et al. (2011) \cite{Hong2011} focused on predicting the number of retweets and analyzed what kinds of tweets attract more retweets.

Besides predicting the number of replies, there is also work on predicting the binary task of whether a Tweet can get a response (e.g., replies or retweets). Yoav Artzi et al.(2012) \cite{Artzi2012} tackle this task mainly based on users' social network and historical influence. 
Rowe et al. (2014) \cite{Rowe2014} further investigated more features that might affect engagement across various social media platforms.

The above-mentioned work mainly focuses on a single post (e.g., a Tweet), which is not part of a longer conversation. On the contrary, we identify the thread-ending posts in the context of a multi-party conversation, which has a complex internal structure and richer context information beyond the text content.

\subsection{Prediction Tasks in online Forums}
Another line of work comes from the domain of online forums. Some studies aim at predicting thread structures. The work proposed by Wang et al. 2011 \cite{Wang2011} approaches this task by detecting initiation-response pairs, which are pairs of utterances that the first part sets up an expectation for the second part. Balali et al. \cite{Balali} followed their work by reconstructing the thread structure, formulating it as a supervised learning task. 

There are other types of tasks in online forums that are relevant to work, including assessing the quality of posts \cite{Siersdorfer2010, Tausczik2011, Ghose2011}, and categorization of post types (e.g., question, solution, spam) \cite{Perumal2016}. 

Although the context and features are relevant, these tasks do not aim at the problem we are solving, to identify thread-ending posts in a conversation. Further investigation needs to be done to identify what types of information could be predictive in our task.


\subsection{Modeling Conversational Interactions}
There has been extensive work on modeling conversational interactions on social media platforms. 
Honeycutt and Herring \cite{Honeycutt2009}  analyzed how to make Twitter more usable as a tool for collaboration. Boyd et al. \cite{Boyd2010} studied how retweeting can be used as a way to converse with others. Ritter et al. \cite{Ritter2010} used unsupervised conversation models to cluster utterances with similar conversational roles. These tasks focus more on the analysis of conversations, rather than predictions.

Recently, researchers begin to study how to automatically generate responses given a conversation history \cite{Shang2015, Sordoni2015, Serban2016}. In this work, we do not aim at generating responses. Rather, we attempt to help users understand whether their posts that are about to be submitted would terminate the thread of the conversation.


\ \\

In summary, our work uses data across different social media platforms and complement the studies mentioned above, but with a unique focus on the effect of a post to the entire conversations. We propose a deep learning model that takes consideration of the content, structure, and context of the entire conversation and predicts the outcome of a single post. This model and our findings could help the aforementioned tasks in literature and help increase user engagement in online conversations.

\section{ConverNet for Thread-ending prediction}
In this section, we propose a specifically designed neural network model that uses information of an entire conversation to predict thread-ending posts. 

We start with a few definitions. A \textbf{post} is a message submitted by a single user, while a \textbf{conversation} is a set of posts concerning a focused topic posted by a group of people. A \textbf{thread} of a conversation is a subset of posts that are organized as a \textit{tree} structure through the reply-to relationships.  
We only focus on threads with at least two posts, as more than one person has to be involved when there is a conversation\footnote{https://www.merriam-webster.com/dictionary/conversation}.
The \textbf{thread-ending post} is a post in a given thread that will not receive any further replies. In this prediction task, we care more about these thread enders than their counterparts, thus we label a thread-ending post as \emph{positive} and the others as \emph{negative}. In this way, the problem is formulated as a binary classification task. Note that we use thread-ending posts as surrogates for conversation killing posts because they are widely available and have explicit labels.

Our model builds on the insight that recurrent neural networks (RNNs) with a specifically designed attention mechanism can have great advantages on dealing with the internal structure of posts in a thread and that additional context information can further boost the classification performance. The reasons are listed as follows.
\begin{itemize}
\item
Posts in a given thread have strong connections between each other. For explicit tree structure and latent connections behind these posts, RNN models are more suitable compared with standard classification methods, e.g., SVMs. Similar to using RNNs to encode sentences in machine translation tasks \cite{Cho2014}, we can also use them to model posts in a thread, which can be used for the further classification task.
\item
Compared with traditional models, deep learning models have an advantage in dealing with data sets of large scale, by training on one batch at a time. This is desirable when we are working on a large amount of user generated content.
\item
We empirically found that context information, e.g., post time and authorship, could greatly complement text information. Therefore we incorporate them into a unified model.
\item
Some conversation threads can be very long, containing tens of posts. Attention mechanisms are usually added over RNNs to solve the problem of long term dependency ~\cite{yang2016hierarchical}. However, because of the fact that different conversations could vary greatly by length, we find that the standard attention mechanism over posts falls short of modeling the longer threads. Therefore, we specifically design an attention mechanism to handle this situation.

\end{itemize}

We propose a recurrent neural network model, called ConverNet, which implements the above design objectives and ideas. 
In the rest of this section, we will give a brief introduction to standard RNNs, followed by the description of our model.
\subsection{Background}
To facilitate readers with different levels of knowledge, we introduce the standard building blocks used by our model here.

\subsubsection{LSTM And BiLSTM}
In ConverNet, we use BiLSTM as the basic building blocks of its architecture. LSTM (Long Short-term Memory) \cite{Hochreiter:1997:LSM:1246443.1246450} units are widely used to build an RNN model and BiLSTM \cite{graves2013hybrid} is one of its extensions.

Below we briefly introduce the basic formulation of a LSTM layer. Given $c_{0}$ as the cell state's initial value and $x_{t}$ as the input of time step $t$, LSTM can be formulated as follows:
\[i_{t} = \sigma(W_{ix}x_{t}+W_{im}m_{t-1}),\]
\[f_{t} = \sigma(W_{fx}x_{t}+W_{fm}m_{t-1}),\]
\[o_{t} = \sigma(W_{ox}x_{t}+W_{om}m_{t-1}),\]
\[c_{t} = f_{t} \odot c_{t-1}+i_{t}\odot h(W_{cx}x_{t}+W_{cm}m_{t-1})\]
\[m_{t} = o_{t} \odot c_{t}\]
and 
\[\sigma (x) = \frac{1}{1+e^{-x}}, h(x) = \frac{1-e^{-2x}}{1+e^{-2x}},\] 
where $m_{t}$ is the output of the LSTM layer in time step t. 

Bidirectional LSTM is an extension of LSTM. It takes the information not only from the forward pass but also from the backward pass. There are two identical independent LSTM kernels in the BiLSTM. At time step $t$, one takes $x_{t}$ and the other takes $x_{T-t}$ as its input, where $T$ is the total time steps needed. The outputs of the two kernels are later aligned according to the time ordinal number and concatenated as the final output of the BiLSTM block.

Bidirectional LSTM can help overcome the problem that the LSTM kernel at the time step $t$ does not know anything about the following inputs sequences. However, it still cannot solve the paradox that the longer the input sequences, the more LSTM layer is about to forget since the hidden unit inside the LSTM is a constant. Therefore, we need to further explore an attention model.

\subsubsection{Layer Normalization}
\textbf{Layer Normalization} technique introduced by Ba et. al \cite{Ba2016} has shown that in the training process of LSTM, layer normalization can have a significant influence on both the training speed and the task performance. In a standard RNN, the summed inputs in the recurrent layer are computed from the current input $x^{t}$ and previous vector of hidden states $h^{t-1}$ which are computed as $a^{t} = W_{hh}h^{t-1}+W_{xh}x^{t}$. The layer normalized recurrent layer re-centers and re-scales using the extra normalization terms:
\[h^{t} = f[\frac{g}{\sigma ^ {t}} \odot (a^{t} - \mu ^{t})+b]\]
\[\mu^{t} = \frac{1}{H}\sum_{i=1}^{H}a_{i}^{t},\sigma^{t} = \sqrt{\frac{1}{H}\sum_{i=1}^{H}(a_{i}^{t}-\mu^{t})^{2}},\]
where $W_{hh}$ are the recurrent hidden to hidden weights and $W_{xh}$ are the bottom up input to hidden weights. In a layer normalized RNN, the normalization terms make it invariant to re-scaling all of the summed inputs to a layer, thus resulting in much more stable hidden-to-hidden dynamics.

\subsection{Context Information for Prediction of Thread-ending Posts}
As mentioned above, in addition to the text content, we investigate a set of context information of a conversation thread that could contribute to the prediction task.
To incorporate them into a unified model, they are implemented as features, which are listed as follows. Generally speaking, there are four types, \emph{length information, sentiment information, background information, and replying property}. 

%
%

\textbf{Length information}

\emph{Post length}: the number of words in a given post.

\emph{Thread length}: the number of posts in a given thread.

\textbf{Sentiment information}

\emph{Sentiment}: the intensity scores of \emph{neutral}, \emph{positive}, and \emph{negative} sentiments of a given post. In this work we simply adopt the scores implemented by \emph{nltk} the VADER Lexicon\cite{Hutto2014}.

\textbf{Background information}

\emph{Conversation background}: the context where the conversation happens. For example, in the movie-dialog data set, this is the information of the movie in which the conversation happens.

\emph{Author features}: background information of the post author. For example, the number of times the author ends a conversation thread in the past.

\textbf{Reply information}

\emph{Replying structure}: basic replying information of a thread, consisting of every post's parent post in this thread.

\emph{Post time}: the post time interval between each post and its previous one, classified into categories of \emph{within an hour}, \emph{within a day}, \emph{within a week} and \emph{no later than a month}. 

\subsection{ConverNet}
\begin{figure*}[t]
\centering
\includegraphics[width=0.86\textwidth]{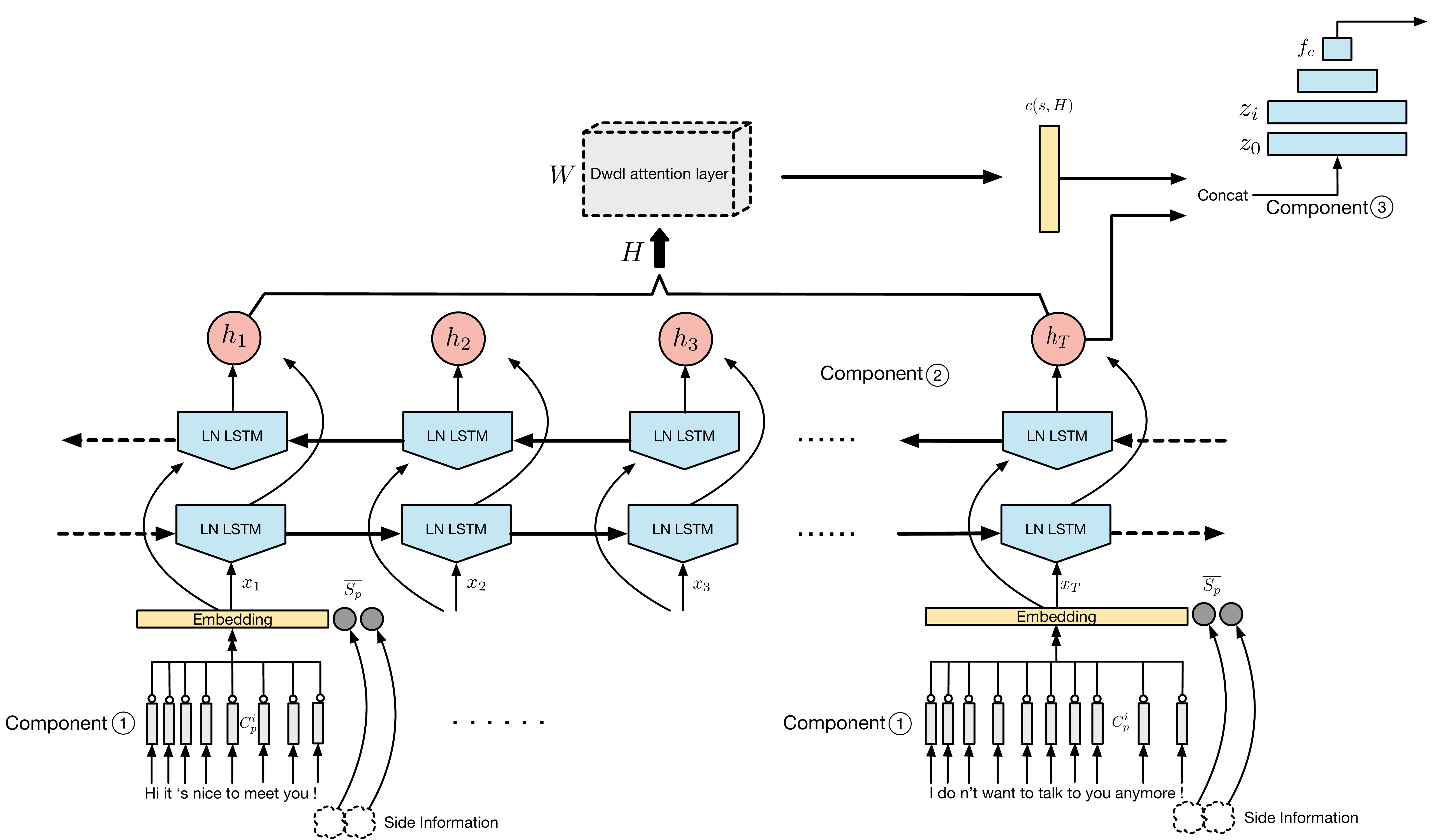}
\caption{A version of the ConverNet designed for the prediction task. Some submodules are numbered and correspondingly detailed in the text. \label{fig:ConverNet}}
\end{figure*}
We now introduce the framework of our proposed neural network model, shown in Figure ~\ref{fig:ConverNet}. For simplicity, we will focus on illustrating three main components in our network.

\ \\
\textbf{(1) Input Processing Component}

The input of ConverNet is a flattened sequence of posts in a thread sorted by their post time, regardless of whether one is replying to the previous one. The replying tree structure is handled along with other context information. 

For the content information, we first use an embedding layer to get $N$ (the total number of posts in a thread) embedding vectors $C_{p}^{i}$ (which represents the embedding for the $i_{th}$ word of the $p_{th}$ post) for each word in a post from a given thread. After that we are going to generate a post embedding vector based on all its words. 
\[x_{t} = [\frac{\sum_{i=1}^{N} C_{t}^{i}}{N};\overline{S_{p}}],\]
where $\overline{S_{p}}$ is the corresponding context information for the $p_{th}$ post in a given thread.
This is done through an average pooling layer on top of the word embedding layer in ConverNet. 

All types of context information are merged with the pooling result by a simple concatenation, composing the input data for LNBiLSTM layer. 

\ \\
\textbf{(2) Encoding Component}

The encoding component consists of a BiLSTM with the layer normalization technique and our proposed Dwdl attention layer, which will be described in details in the next subsection. Firstly, the LNBiLSTM encodes the $x_{i}$ in the following way,
\begin{small}
\[i_t = \sigma_i(LN(x_t W_{xi}; \alpha_{xi}, \beta_{xi})
               + LN(h_{t-1} W_{hi}; \alpha_{hi}, \beta_{hi})
               + w_{ci} \odot c_{t-1} + b_i)\]
\[f_t = \sigma_f(LN(x_t W_{xf}; \alpha_{xf}, \beta_{xf})
               + LN(h_{t-1} W_{hf}; \alpha_{hf}, \beta_{hf})
               + w_{cf} \odot c_{t-1} + b_f)\]
\[c_t = f_t \odot c_{t - 1}
               + i_t \odot \sigma_c(LN(x_t W_{xc}; \alpha_{xc}, \beta_{xc})
               + h_{t-1} W_{hc}; \alpha_{hc}, \beta_{hc})
               + b_c)\]
\[o_t = \sigma_o(LN(x_t W_{xo}; \alpha_{xo}, \beta_{xo})
               + LN(h_{t-1} W_{ho}; \alpha_{ho}, \beta_{ho})
               + w_{co} \odot c_t + b_o)\]
\[h_t = o_t \odot \sigma_h(LN(c_t; \alpha_c, \beta_c)),\]
where
\[LN(z;\alpha, \beta) = \frac{(z-\mu)}{\sigma} \odot \alpha + \beta\]
\end{small}

The parameters are as follows:

For the input gate: $i_t: W_{xi}, W_{hi}, w_{ci}, b_i$, and $\sigma_i$.

For the forget gate: $f_t: W_{xf}, W_{hf}, w_{cf}, b_f$, and $\sigma_f$.

For the cell computation: $c_t: W_{xc},W_{hc},b_c$, and $\sigma_c$.

For the output gate: $o_t: W_{xo},W_{ho},w_{co},b_o$, and $\sigma_o.$

The post to be predicted is positioned at the end of the input sequence. However, instead of using only the final output of the LNBiLSTM layer. $h_T$, we take full use of all the outputs $\{h_t\}$ from all time steps. The Dwdl attention layer functions to further encode the output sequence $[h_{t}]$ of LNBiLSTM into a vector that has the same dimension as the hidden units in LNBiLSTM, where $[h_{t}]$ is a matrix \textbf{vertically} stacked by LNBiLSTM output $h_{t}$ at every time step time. 

In the end, there is a merge operation implemented to combine the result of the attention layer and the LNBiLSTM's last output vector. 
\[z_{0} = tanh([c(s, H); h_{T}]\cdot W)\]
In this way, theoretically, the performance of adding the attention layer will not be any worse than a single LNBiLSTM kernel. As for the merge operation, we implement it using \emph{concatenation}.

\ \\
\textbf{(3) Decoding Component}

After getting the result from the encoding component, the decoding component focuses on the final classification task. It consists of several MLP layers, followed by the final layer that has only one output unit, predicting whether the given post is a conversation en or not: 
\[z_{i} = Relu(z_{i-1}\cdot W+b)\]
\[\hat{y} = Sigmoid(z_{n}\cdot W+b)\]

All MLP layers are followed by batch normalization layers\cite{Ioffe2015}. All layers are activated with ReLU \cite{Glorot2011}, with the exception of the last one, which is activated with \emph{Sigmoid} function. This guarantees that the final output is either 0 or 1.

\subsection{Dwdl Attention Layer}
The motivation to use an attention layer is to take full usage of the information generated by the LSTM kernel. However, one difficulty of our task is that the length of threads ranges from a considerable scale. The standard attention mechanism learns attention weights uniformly over posts. That is, the $i$-th post in different threads always receive the same attention. Unfortunately, this assumption does not hold in reality, as the learned weights do not fit universally to posts in threads of various lengths. 

As a solution, one may want to apply different attention weights for each thread. This leads to another problem by introducing a large number of parameters to be learned.

To resolve both issues, we propose an attention mechanism that applies \textbf{d}ifferent attention \textbf{w}eights for \textbf{d}ifferent \textbf{l}engths of input \textbf{(Dwdl)}, while weights are shared among threads of identical length. 
In this way, the attention mechanism thus outputs the result $c(s, H)$, which is,
\[c(s, H) = \frac{\sum_{k=1}^{s}[exp(w_{ks}) \cdot h_{k}]}{\sum_{k=1}^{s}exp(w_{ks})},\]
where $H$ is the output sequence from the LNBiLSTM layer with length $s$. And $W$ is the attention weight matrix that will be learned.

Despite its simplicity, we found that it solves not only the problem that it is hard to learn thread representations when they vary greatly in length, but also avoid to introduce too many parameters to be learned. The design of the Dwdl attention layer is a major innovation of the proposed ConverNet model in the context of deep learning architectures. 
%

\subsection{Loss function}
We use binary cross entropy loss to train our model.
The objective is to minimize the loss function:
\[\mathcal{L} = - \sum_{i} [g_{i} \\log(\hat{y}_{i}) + (1 - g_{i}) \\log(1 - \hat{y}_{i})],\]
where $\hat{y}_{i}$ is the predicted probability of the $i$-th conversation, and $g_{i}$ is the ground-truth label.

\section{EXPERIMENT SETUP}
We present empirical experiments that compare our model with various alternative approaches on two public datasets. The following experiments are aimed to demonstrate the effectiveness of ConverNet in general and the effectiveness of different kinds of content and context information in predicting thread-ending posts.

\subsection{Data Sets}
We accomplish this task on two representative datasets. One contains threads of Reddit posts and comments, which is extracted from Reddit.com, one of the largest online forums that cover a variety of topics. This dataset is representative for online conversations. The other is a collection of conversations extracted from movie scripts. We include this dataset because movie dialogs are closer to offline, everyday conversations, which would be a good reference for understanding the properties of online conversations. 
The statistics of these datasets are listed in Table~\ref{tab:statistics_data}.

\begin{table}[h!]
\caption{Statistics of each data set.}
\label{tab:statistics_data}
\small
\begin{tabular}{  c|c|c }
\hline
Properties & Reddit-Threads & Movie-Dialogs\\
\hline
Threads & 83,097 & 100,000\\
\hline
Vocabulary & 29,729 & 107,354\\
\hline
Max post len. & 673 & 2689\\
\hline
Avg. post len. & 13.02 words & 43.83 words\\
\hline
\# train threads & 63,097 & 80,000\\
\hline
\# val threads & 10,000 & 10,000\\
\hline
\# test threads & 10,000 & 10,000\\
\hline
\end{tabular}
\end{table}

\textbf{Reddit-Threads Data Set}

\emph{Source.} This data set is generated based on the public Reddit-Comments data set provided by Reddit user Stuck\_In\_the\_Matrix \cite{Reddit}. The original data set consists of all of the posts and comments available on Reddit since early 2006. In our experiment, we focus on the threads in the political domain (from 2007.8$\sim$2009.8), a major topic of interest on Reddit. 

\begin{figure}[t]
\includegraphics[width=0.47\textwidth]{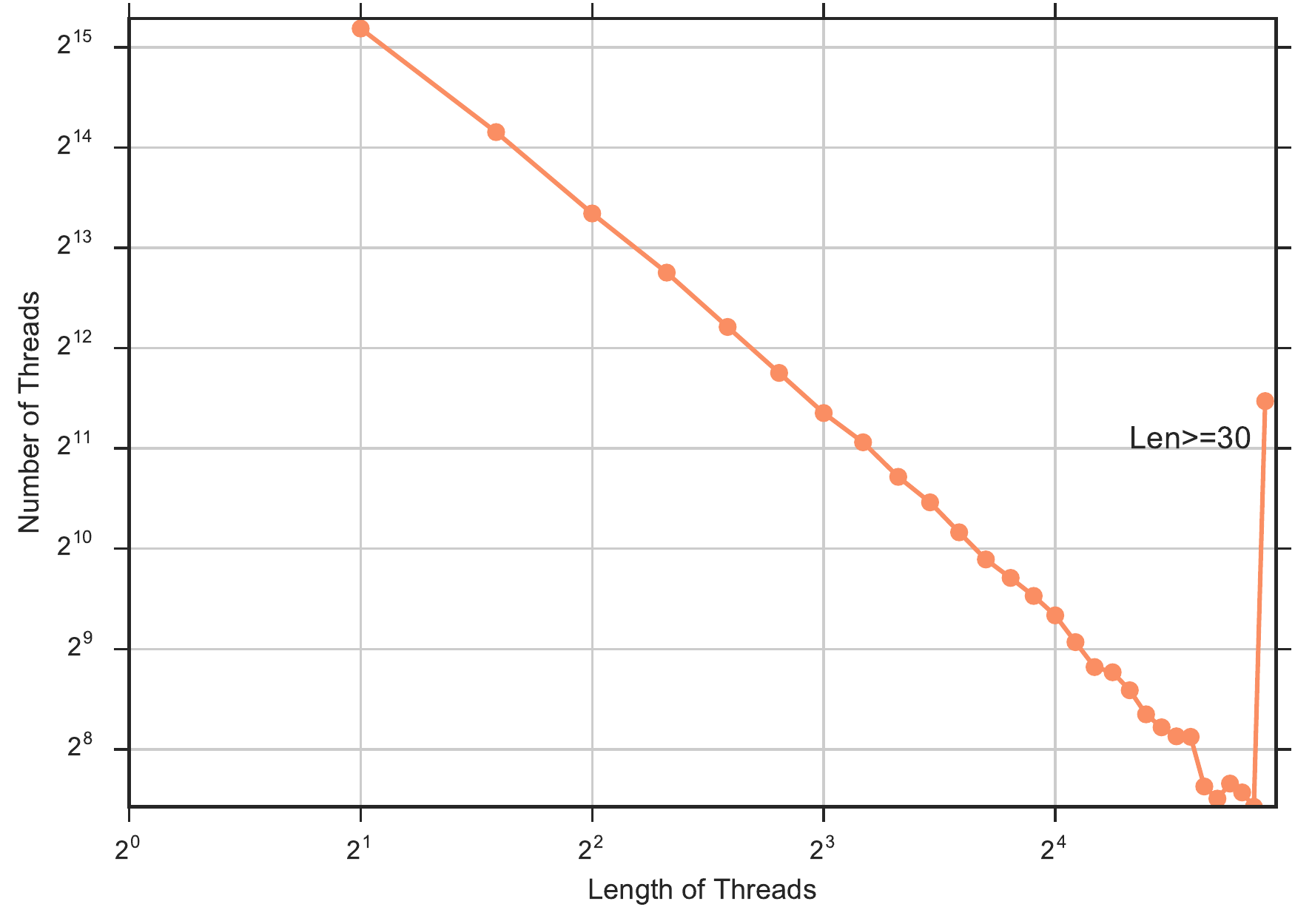}
\caption{The length distribution of threads in Reddit-Threads data set.\label{fig:len_distr_reddit}}
\end{figure}

\emph{Processing.} By utilizing the \textit{parent} post information provided by Reddit, we recover the tree structure of each thread, where leaf nodes are considered as thread-ending posts. As mentioned before, we only focus on threads with more than one post -- a thread with only one post is not a conversation. The length distribution of threads is shown in Figure~\ref{fig:len_distr_reddit}, which generally follows the power law distribution. 

\emph{Prediction Task.} With these tree-structured threads, our prediction task is equivalent to predicting whether a given node is a leaf node or not. Note that in a Reddit Thread, there might be more than two actors (authors) engaged in a conversation. 

\textbf{Movie-Dialogs Corpus}

\begin{figure}[t]
\includegraphics[width=0.47\textwidth]{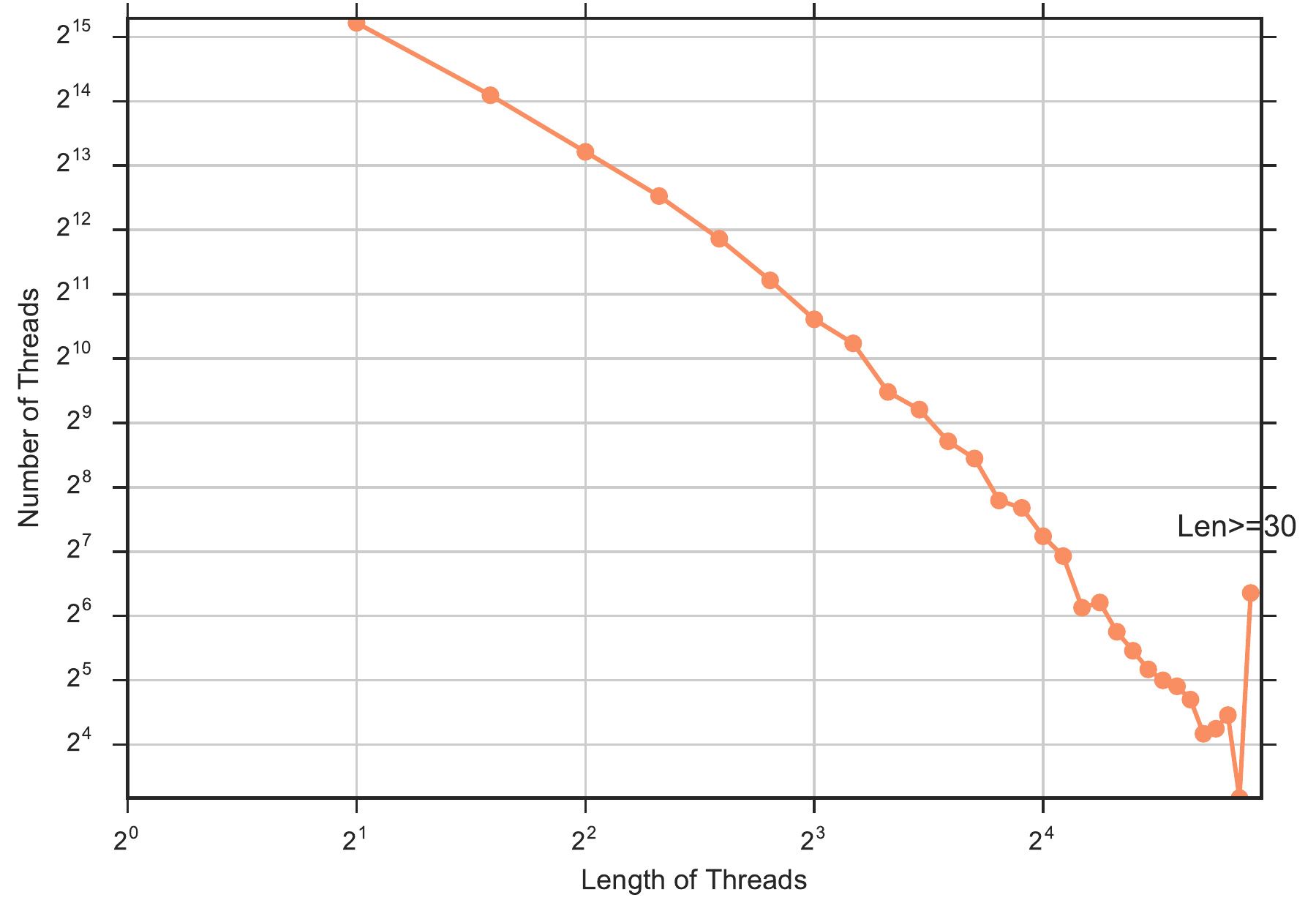}
\caption{The length distribution of threads in Movie-Dialogs data set.\label{fig:len_distr_movie}}
\end{figure}

\emph{Sources.} We use the Cornell Movie-Dialogs Corpus, which is widely used for text generation tasks \cite{Danescu-Niculescu-Mizil+Lee:11a}. It contains more everyday words and involves in total 617 movies with 10,292 movie characters. Compared with the Reddit-Threads data set, the posts (or sentences) are simpler, shorter, and more formal. A major difference is that every dialog happens between two speakers, so the number of users in threads is a constant. The length distribution of threads is shown in Figure~\ref{fig:len_distr_movie}.

\emph{Prediction Task.} We treat every movie dialog also as a chatting thread with two participators. These chatting threads are organized as a sequence of ``posts'' (sentences) instead of a tree structure. As a result, only the last post (sentence) ends the conversation. 

\textbf{Sampling and Other Processing}

In both datasets, we randomly sample one post (sentence) from each thread (dialog) and predict whether it is a thread-ending post. We call these posts \textbf{target posts}. Since the information after the target post will reveal the ground truth for the prediction task, all posts after the target post must be omitted before putting into the model. 
The Reddit-threads data set are split into train, validation, and test set according to submission time of the first post in each thread. The first 80,000 threads are assigned to the training set and the rest 20,000 are equally separated into validation and test set. For the movie data set, we do a random permutation of all the threads, and assign them into train/val/test set as Table~\ref{tab:statistics_data} shows.


\subsection{Metrics}
Since the label distribution of our binary classification task is skewed, we adopt metrics in addition to the commonly used \textbf{accuracy}, including \emph{AUC} and \emph{MAP} (mean-average precision), because in reality it is more important to make sure the top-ranked posts are truely thread-ending posts (so that notices can be sent). To achieve high scores when evaluated by these additional metrics, both precision and recall are important.



\subsection{Competing methods}
We compare ConverNet with a handful of baseline methods in two categories: conventional machine learning methods and alternative deep learning methods.

\subsubsection{Conventional Baselines}\textbf{ }\\
\textbf{SVM+/-[features].} As a conventional method that has demonstrated superior performance in many kinds of classification tasks, SVM can sometimes achieve comparable performance to the deep learning methods. Besides, it plays a key role in deciding which kind of hand-crafted features are helpful in our prediction task. Therefore, we include all kinds of features potentially related to this problem across domains, and use a linear kernel SVM implemented by \emph{sklearn} for classification. 

In addition to all the features mentioned in the method section, we include additional features extracted from text content. They include word unigrams, bigrams, trigrams, and post embeddings. Specifically, post embeddings are the averages of word vectors in a post. The word vectors are generated by word2vec~\cite{mikolov2013efficient} skip-gram and CBOW models. We concatenate all the features from all the posts in a thread. When a method is named \textbf{SVM+[features]}, it refers to an SVM model trained using only the corresponding set of features. When a method is named \textbf{SVM-[features]}, it refers to an SVM model that uses all but the corresponding set of features. 



\subsubsection{Deep learning baselines}\textbf{ }\\
\textbf{BiLSTM}, \textbf{LNBiLSTM}, \textbf{Parallel LNBiLSTM}. We use the bi-directional LSTM (\textbf{BiLSTM}) model that is widely used for classification tasks. Considering the recent success of layer normalization, we include the bi-directional LSTM with layer normalization (\textbf{LNBiLSTM}) as a baseline. We also stack multiple LNBiLSTM to learn deeper representations (\textbf{Stacked LNBiLSTM}).

\textbf{LNBiLSTM+features} and \textbf{LNBiLSTM+features+SA}. One major innovation of ConverNet is the newly designed Dwdl attention mechanism (see Section 3.4) to handle context information in a thread. For comparison, we also add context information to LNBiLSTM using several ways. \textbf{LNBiLSTM+features} simply concatenate the representation output by LNBiLSTM and features extracted from context information. \textbf{LNBiLSTM+features+SA} applies a standard attention over the output of the hidden state by LNBiLSTM. For all LNBiLSTM-related models, LSTMs are stacked using horizontal or vertical ways. But the stacking times can vary from model to model.


\subsection{Training Details}

\begin{table*}[t]
\caption{Performance of competing methods: LNBiLSTM+All features+Dwdl attention achieves top performance. }
\label{tab:exp_results}
\begin{center}
\setlength{\tabcolsep}{6pt}
\renewcommand{\arraystretch}{1.1}
\footnotesize
\begin{tabular}{c|c|c|c||c|c|c}  
\hline
 & \multicolumn{3}{c||}{Reddit-Threads Data Set} &  \multicolumn{3}{|c}{Movie-Dialogs Data Set}\\
\hline
Method & Accuracy  & AUC & MAP & Accuracy & AUC  & MAP\\
\hline 
SVM-Text content(Embedding, N-grams) & $75.95\scriptstyle\diamond\diamond$ & $81.26\scriptstyle\diamond\diamond\diamond$ &
$68.84\scriptstyle\diamond\diamond\diamond$ & $74.57\scriptstyle\diamond\diamond\diamond$ & $83.12\scriptstyle\diamond\diamond\diamond$ &
$64.97\scriptstyle\diamond\diamond\diamond$\\
SVM-Lengths info & $76.45$ & $83.05$ & $72.31$ & $75.70$ & $84.67$ & $69.50$\\
SVM-Background info & $-$ & $-$ & $-$ & $75.43\scriptstyle\diamond$ &
$84.56$ &
$69.09\scriptstyle\diamond$\\
SVM-Post time & $75.55\scriptstyle\diamond\diamond\diamond$ & 
$81.36\scriptstyle\diamond\diamond\diamond$ &
$69.85\scriptstyle\diamond\diamond\diamond$ & 
$-$ & $-$ & $-$\\
SVM-Replying structures & $76.15$ & $83.13$ & $72.67$ & $-$ & $-$ & $-$\\
SVM-Sentiment & $76.31$ & $83.06$ & $72.84$ & $75.60$ & $84.59$ & $69.59$\\
SVM+All features & $76.39$ & $83.30$ & $72.60$ &  $75.84$ & $84.67$ & $69.63$\\
\hline
BiLSTM+Text content (only the target post) &  $60.80\scriptstyle^{\star\star\star}_{\diamond\diamond\diamond}$ & 
$64.36\scriptstyle^{\star\star\star}_{\diamond\diamond\diamond}$ &
$58.20\scriptstyle^{\star\star\star}_{\diamond\diamond\diamond}$ &
$61.62\scriptstyle^{\star\star\star}_{\diamond\diamond\diamond}$ & 
$61.40\scriptstyle^{\star\star\star}_{\diamond\diamond\diamond}$ & $50.30\scriptstyle^{\star\star\star}_{\diamond\diamond\diamond}$\\
\hline
BiLSTM+Text content &  $76.02\scriptstyle\star\star\star$ &
$83.42\scriptstyle\star\star\star$ &
$73.33$ & 
$76.26\scriptstyle\star\star\star$ & 
$85.22\scriptstyle\star\star\star$ & $70.63\scriptstyle\star\star\star$\\
LNBiLSTM+Text content &  $76.59\scriptstyle\star\star\star$ & 
$84.22\scriptstyle\star\star\star$ &
$74.07\scriptstyle\star\star\star$ & $76.75\scriptstyle\star\star$ & 
$85.55\scriptstyle\star\star\star$ & $70.85\scriptstyle\star\star\star$\\
Stacked LNBiLSTM+Text content & $76.42\scriptstyle\star\star\star$ & 
$84.46\scriptstyle\star\star\star$ & $74.44\scriptstyle\star\star\star$ & $76.98\scriptstyle\star$ & 
$85.87\scriptstyle\star\star$ & $71.67\scriptstyle\star\star$\\
\hline
LNBiLSTM+All features & $78.05$ & $85.91$ & $77.39$ & $77.51$ & $86.47$ & $72.95$\\
LNBiLSTM+All features+Standard attention & $78.05\scriptstyle\diamond\diamond\diamond$ & 
$85.97\scriptstyle\diamond\diamond\diamond$ & $77.70\scriptstyle\diamond\diamond\diamond$ & $77.45\scriptstyle\diamond\diamond\diamond$ & 
$86.32\scriptstyle\diamond\diamond\diamond$ &
$72.63\scriptstyle\diamond\diamond\diamond$\\
\hline
\textbf{ConverNet} &  $78.27\scriptstyle\diamond\diamond\diamond$ & 
$86.22\scriptstyle^{\star\star}_{\diamond\diamond\diamond}$ &				
$78.21\scriptstyle^{\star}_{\diamond\diamond\diamond}$ &  $78.04\scriptstyle^{\star\star}_{\diamond\diamond\diamond}$ & 
$86.82\scriptstyle^{\star\star\star}_{\diamond\diamond\diamond}$ &
$73.76\scriptstyle^{\star\star\star}_{\diamond\diamond\diamond}$\\
\hline
\end{tabular}
\flushleft All numbers are in percentage. $\scriptstyle  \star(\star\star,\star\star\star)$ indicates that one method is statistically significantly better or worse than \textit{LNBiLSTM+All features+Standard attention} (which is in general the best configuration among all non-ConverNet, LSTM-related models) according to Random Permutation Test  \cite{fisher193577m} at the significance level of 0.05(0.01,0.001). And $\scriptstyle\diamond(\diamond\diamond,\diamond\diamond\diamond)$ indicates one method is statistically significantly better or worse than \textit{SVM+All features} at also the significance level of 0.05(0.01,0.001). Results are indicated by ``-'' if a feature category is not available for a particular data set.
\end{center}
\renewcommand{\arraystretch}{1}
\end{table*}

All hyper-parameters are tuned to obtain the best performance of AUC score on the validation set. For LSTM-related methods, the candidate word embedding sizes are set as $\{16,$ $32,$ $64,$ $128,$ $256\}$ and the candidate numbers of hidden/cell units in the LSTM-related layer are $\{16,$ $32,$ $64,$ $128\}$. The vertically and horizontally LSTM stacking candidate number is chosen from $\{1, 2, 3\}$. The embedding size for context information is selected from $\{2, 4, 8, 16, 32, 64\}$. The initial learning rate is selected from $\{10^{-1}, 10^{-2}, ..., 10^{-5}\}$. For SVM, the candidate embedding size is from $\{50,$ $100,$ $200,$ $500\}$ and the relaxing parameter C for SVM model is chosen from $\{10^3, 10^2, ..., 10^{-3}\}$. 

We initialize parameters in neural networks using a zero-mean Gaussian with standard deviation selected from $\{0.01,$ $0.05,$ $0.1,$ $0.2\}$.
For parallel network models, the number of stacking layers is selected from $\{2,$ $3,...,$ $5\}$. 
All deep learning models are optimized by RmsProp~\cite{Tieleman2012}. 
We stop training when the performance converges on the validation set.
\section{experiment results}
\subsection{Overall Performance}

The overall performance of all competing methods are shown in Table~\ref{tab:exp_results}. 
The proposed method ConverNet outperforms all competing methods in all three metrics, \emph{AUC}, \emph{Accuracy}, and \emph{MAP}. The improvements are all statistically significant except for one case (accuracy on Reddit-Threads, where LNBiLSTM+All features already performs very well). This empirically confirms that a well designed deep learning model can achieve the best result in predicting thread-ending posts in online conversations.

Comparing different versions of SVM models, the content of the thread and the target post appears to be the most important. When content features are included, there is a 5\% improvement in MAP on Reddit dataset (0.688 -> 0.726) and 7\% improvement on the Movie-dialog dataset (0.650 -> 0.696). Certain context information is also useful on top of textual features, especially the time of the posts in Reddit threads (0.699 -> 0.726). A more detailed comparison of the features is deferred to Section 6. Consistent conclusion can be made by comparing ConverNet with the best deep learning baseline that is purely based on content information. Overall, comparing ConverNet to the best performing SVM baseline, there is another 8\% improvement on Reddit-Threads (0.726 -> 0.782) and 6\% improvement on Movie-Dialogs (0.696 -> 0.737).   

When using the standard attention layer to handle context information, the deep learning model does not perform better than the model based on content only, sometimes even less effective. This is possibly due to the large variation of the thread length in both data sets. Threads with fewer posts could have a totally different attention distribution from those with more posts. 
With the newly designed Dwrl attention layer, ConverNet is able to significantly outperform the content based models. 

The experiment results of using different LSTM-related kernels also prove that BiLSTM with layer normalization can make a significant improvement, while repeatedly stacking this layer can also slightly enhance the prediction result. 

It is interesting to note that a deep learning model learned only based on the content of the target post (instead of the whole thread), \textbf{BiLSTM+Text content (only the target post)}, performs significantly worse even compared to the same model that considers the content of both the target post and other posts in the thread (with no context information). This assures that information in the whole thread is important for predicting whether a post will be replied to, which again distinguishes our problem setting with existing work that predicts retweets. Indeed, whether a post will carry on a conversation highly depends on whether what it says is relevant to the topic of the discussion. 


\subsection{Training Time Analysis}
In order to measure the training speed of each model, we train all deep learning methods on a server with a single TITAN X GPU (12 GB GDDR5X, Graphics Card Power of 250 W).

For all deep learning methods, the numbers of epochs required for training are all quite close, which is around 15 epochs averaged over all data sets.

Comparing the training time per epoch, stacking more LNBiLSTM will decrease the time efficiency. And the basic BiLSTM model takes around 55 seconds per epoch. Using layer normalization will slightly reduce training time (2~3 seconds). Adding attention layer increases the training time. Our ConverNet model consumes around 60 seconds per epoch. The total training time of ConverNet model is around 900 seconds with about 18 training epochs.
\section{Discussion}

The overall performances of the competing algorithms have demonstrated that thread-ending posts are predictable through integrating content and rich context information, and through a carefully designed deep learning architecture. Beyond the numbers, we are also interested in the implications of the experiment on how to avoid to be a conversation killer. We approach this by a more detailed analysis of the features and interpretation of the model results. 

\subsection{Feature Analysis}

Based on the features extracted for SVM, we first conduct a simple correlation analysis to understand what features of content and context are positively or negatively correlated with the outcome, whether a post in a thread-ending post or not. Correlations of some selected features (measured in Pearson's coefficient) and the outcome label are shown in Table~\ref{tab:discussion}. 

\begin{table}[h!]
\caption{Correlation of features to ending a thread.}
\label{tab:discussion}
\small
\begin{tabular}{  c|c|c }
\hline
 & \multicolumn{1}{c|}{Reddit-Threads} &  \multicolumn{1}{|c}{Movie-Dialogs}\\
\hline
Features & Correlation & Correlation\\
\hline
Word Embeddings(-) & \multicolumn{2}{|c}{`Mr.',`Mrs.',`like',`talked',`heard',`seen',`care'}\\
\hline
Word Embeddings(+) & \multicolumn{2}{|c}{`ass',`but',`YOU'}\\
\hline
Length of Thread & + & + \\
\hline
Length of Post & - & + \\
\hline
Post Time Difference & + & x \\
\hline
Positive Sentiment Score & + & -\\
\hline
Negative Sentiment Score & - & +\\
\hline
\end{tabular}
\flushleft For a feature, we report the sign of Pearson's coefficient to class label. The mark ``x'' means that feature does not exist in the dataset, ``/'' means the coefficient is trivial, + and - represent significantly (p < 0.01) positive (likely to end thread) or negative correlation (unlikely to end thread).
\end{table}

Embeddings of certain words are identified as significantly correlated to thread-ending posts, either positively or negatively (positive words mean higher probabilities to cause endings, negative words otherwise). We see that the most correlated words are likely to be sentimental or particular expressions instead of topical words. For example, polite addresses like 'Mr.' and 'Mrs' will more successfully lead to further communications. Key Words indicating inclines of sharing experiences like 'seen', 'heard' will also draw the attention of other users. However, insulting words like 'ass' or words with an intense sentiment like 'YOU' will be more likely to end a conversation. 

Length of a thread is positively correlated with thread-ending posts, indicating that being the first a few posts in a conversation is more likely to be replied to, and when the conversations is already lengthy, it is less likely to prolong the discussion. Post time (time since the previous post) is also positively correlated to the outcome, indicating that the longer one waits to reply to a thread, the more likely they will never be replied to. 

These findings are quite intuitive and pretty much consistent on two datasets. There are other features that are more intriguing. For example, in online conversations (Reddit threads), the more words a post has, the more likely that it brings in a reply. In everyday conversations (movie dialogs), a long speech does not necessarily bring in responses. Saying too much might result in a silence. Perhaps reading (a forum post) is indeed more efficient than listening? In movie dialogs, a post that is more positive is less likely to end a conversation and negative sentiment is the contrary, which is consistent to our general intuition (being more polite to get people respond to you). Interestingly, these correlations are the opposite in Reddit threads. Considering the unique nature of online forums and the topic (politics), it is perhaps not too surprising. On one hand, political discussions in online forums are known to be intense and controversial, where an attacking or uncivil (usually with extremely negative sentiment) post is likely to bring another negative reply \cite{cheng2017anyone}. On the other hand, many threads in online forums are asking questions, and when a satisfactory answer is provided, these threads are usually ended with a short post of a simple appreciation. Therefore, a positive sentiment is linked to thread-ending. Apparently, in these two cases, ending a thread politely is not a bad thing, and prolonging an uncivil discussion is much more undesirable. This does intrigue us to rethink about the difference between thread-ending posts and conversation killers, and between conversation killers and killers of a ``good conversation.''  

Given the correlation analysis of individual features, we are also interested to know how the features work together (see Table~\ref{tab:exp_results}). Because different types of content and context features can be highly correlated, when they are feed into SVM, the signs and coefficients of the features may or may not be consistent with the correlations.  

In the Movie-Dialog data set, apart from the most predictive content features, movie background features (such as the theme of the movie) also contribute significantly. On the contrary, sentiment and length features are less useful on top of content features, which maybe because such information might have already been captured by the content features. The background feature represents the circumstances under which conversations happen. One may also utilize such background information for analyzing online forums if it is available (e.g., politics vs. entertainment).  

In the Reddit-Threads data set, post time brings in an significant improvement, while replying structures and sentiment scores can only slightly improve on top of other features. This is because sentiment scores and replying structures may have become redundant when other kinds of features are already used. For example, SVMs might have already learned the sentiment feature from the text content. However, context information like post time is relatively orthogonal to the other types of features, thus bringing in a more noticeable improvement. 


\subsection{What ConverNet Learns}
To gain a better understanding of the behavior of ConverNet, we manually analyzed cases where ConverNet performs better than SVMs. These cases can be put into the following categories.

\emph{Posts with an intense tone.}

\begin{figure}[h]
\centering
\includegraphics[width=0.45\textwidth]{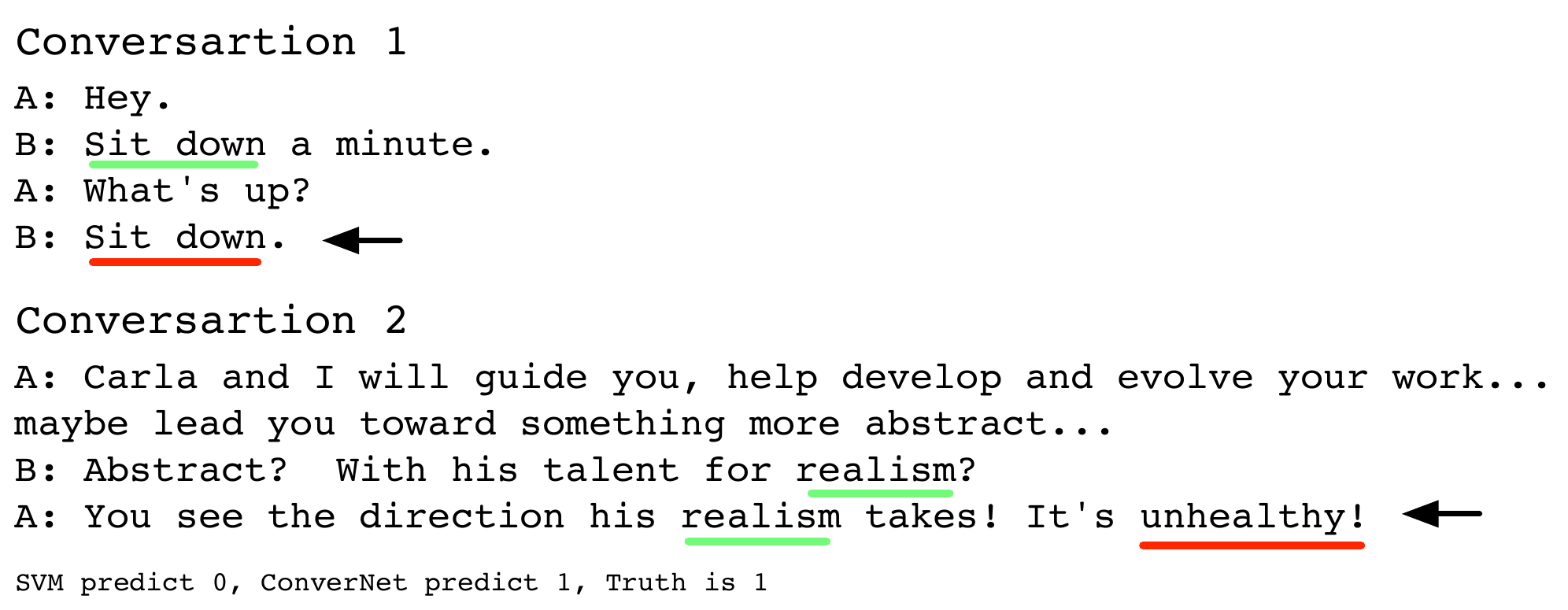}
\caption{ConverNet performs better than SVM on thread-ending posts with an intense tone.}
\label{fig:post_intense}
\end{figure}

Empirically speaking, a post with an intense tone tends to create a serious chatting atmosphere, where other conversation participants may get nervous or shocked to say anything, thus increasing the possibility to end a conversation. It is hard for SVMs to identify these cases even with the help of sentiment analysis, while ConverNet can have a significantly higher chance to give the right prediction. An example is shown in Figure~\ref{fig:post_intense}. \textit{Sit} and \textit{down} are both everyday words. Sentiment lexicons might fail to detect any sentiments, but we can sense a commanding tone of the expression, especially for the last utterance. Such subtlety can be detected by neural network's recurrent mechanism and the attention mechanism.

\emph{Posts with an inquiry tone.}

\begin{figure}[h]
\centering
\includegraphics[width=0.45\textwidth]{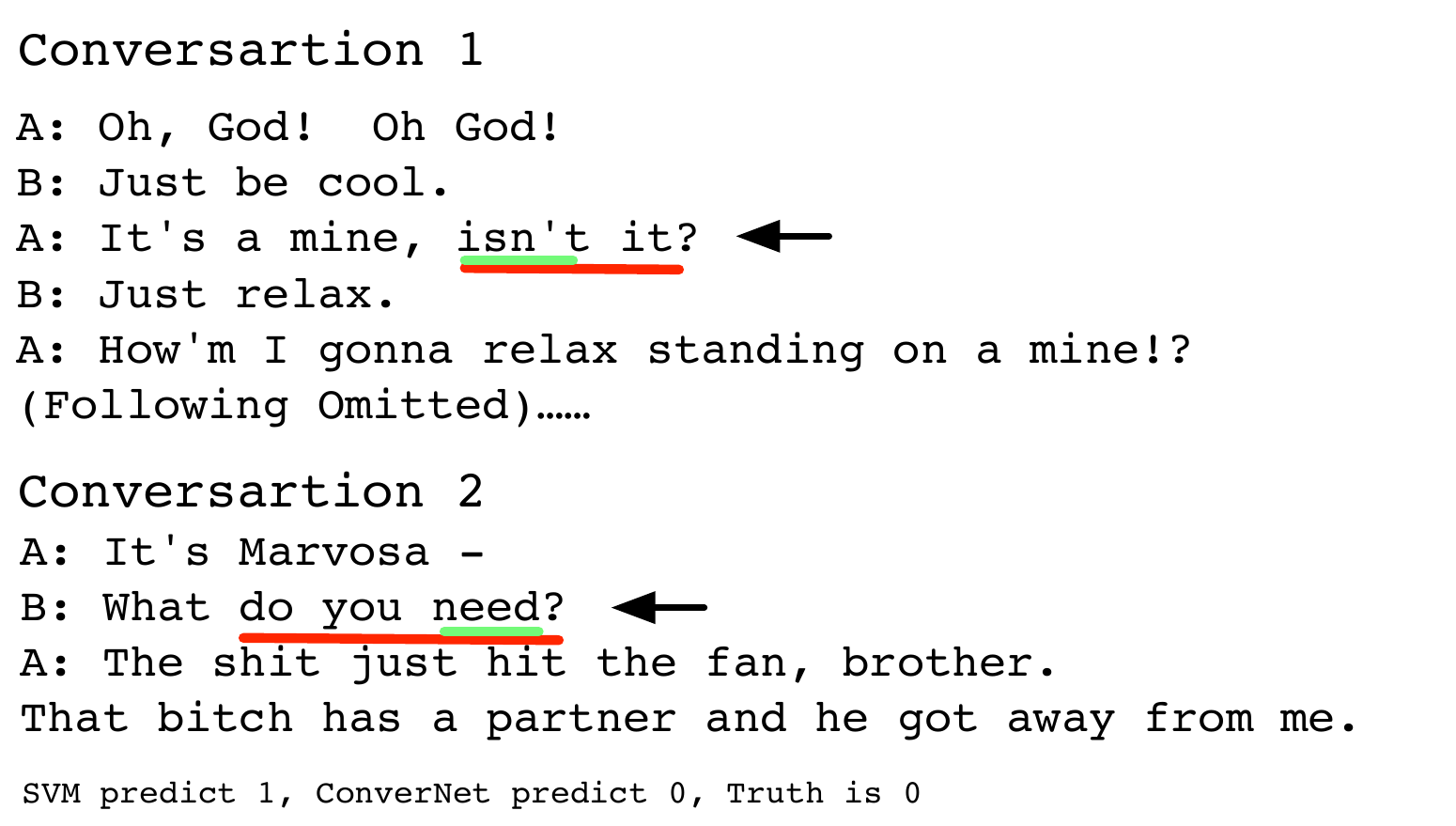}
\caption{ConverNet performs better than SVM on thread-ending posts with an asking tone.}
\label{fig:post_asking}
\end{figure}

If someone is asking for other people's opinions or proposing other questions, the conversation may have a much higher possibility to carry on. When these posts are the targets, ConverNet is more likely to detect these questions and give a correct prediction. Some examples are shown in the Figure \ref{fig:post_asking}.

\emph{Posts with a vague tone.}

\begin{figure}[h]
\centering
\includegraphics[width=0.45\textwidth]{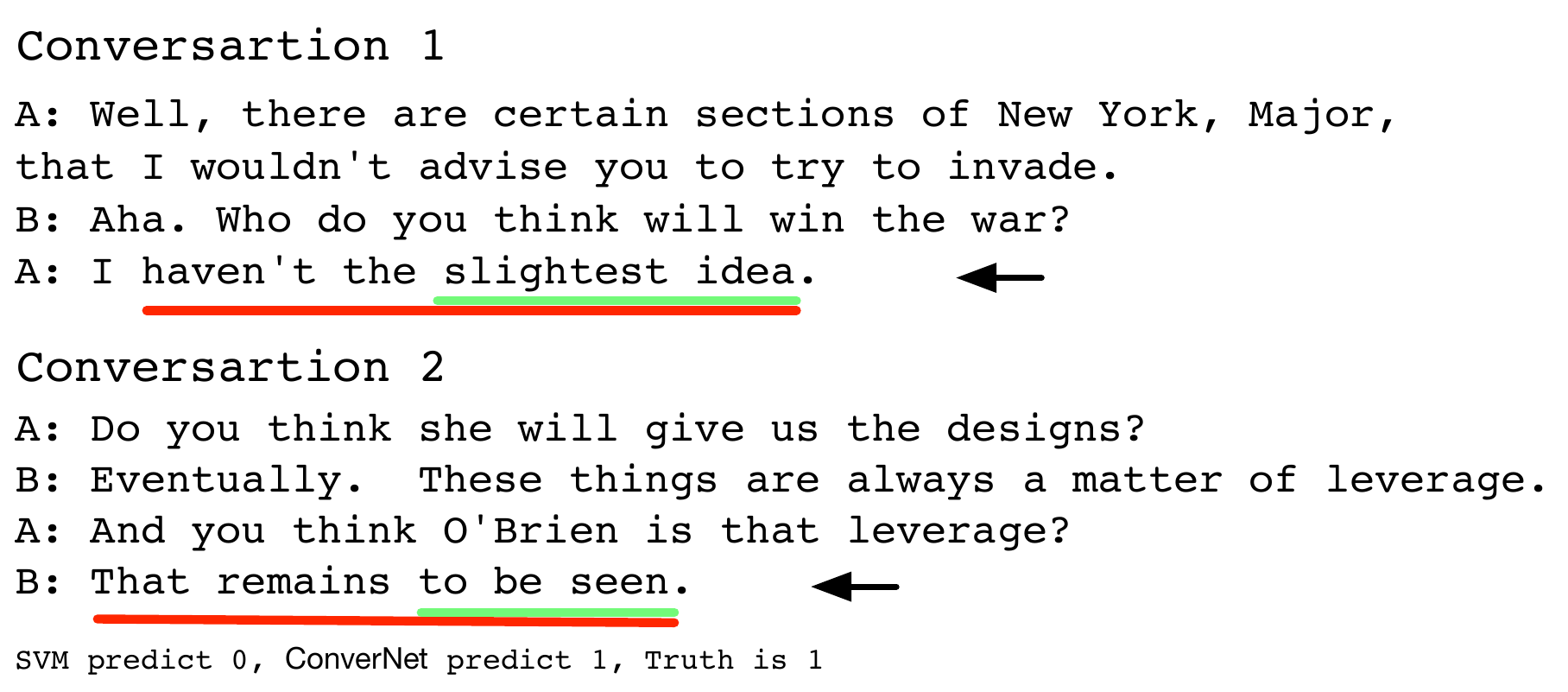}
\caption{ConverNet performs better than SVM on thread-ending posts with a vague tone.}
\label{fig:post_vague}
\end{figure}

We find that a large number of thread-ending posts carry a vague tone, with examples listed in the Figure \ref{fig:post_vague}. These posts convey ambiguous meanings, giving no direct response to the questions raised by others before. This kind of vague tone can make other participants think that the speaker is not interested, or is not giving enough attention, which in turn makes the conversation stop. ConverNet's recurrent mechanism and attention mechanism can help better extract these ambiguous words from given comments, thus yielding a higher precision.

Based on the analysis of features and the interpretation of the models, if advices have to be given to avoid being a conversation killer, some general implications may be: keep to the point (content), act fast (post time and thread length), be elaborative (post length), be positive (sentiment), and pay attention to your tone (deep patterns in language). 

%

%
\section{Conclusion}
How to improve the quality of conversations and engage user participation in online communities is a critical problem that may be relevant to every Internet user. Our work focuses on a novel data mining problem, to identify what type of posts are likely to end a thread of conversation online. We find that while a standard SVM can effectively identify useful signals from both content and context information that are predictive for thread-ending posts, a carefully designed recurrent neural network model, ConverNet, is able to maximize the predictive power of these signals. ConverNet outperforms all the competing baselines in data sets from two representative domains. 
The results of ConverNet also provides practical implications for improving the quality of online conversations. Our work opens up interesting directions towards understanding the quality of online conversations and increasing user engagement, and towards a deeper understanding of the functionality of language in a conversation. 

\subsection*{Acknowledgment} This work was supported in part by the 973 program (2015CB352302), NSFC (U1611461) and key program of Zhejiang Province (2015C01027), and was partially supported by the National Science Foundation under grant numbers IIS-1054199, IIS-1633370, and SES-1131500.

\balance 
\bibliographystyle{ACM-Reference-Format}

\end{document}